# Validation of AI-Based 3D Human Pose Estimation in a Cyber-Physical Environment

1st Lisa Marie Otto
*Fachgebiet Kraftfahrzeuge*
*TU Berlin*
Berlin, Germany
0009-0002-5891-7316

2nd Michael Kaiser
*Fachgebiet Kraftfahrzeuge*
*TU Berlin*
Berlin, Germany
0009-0005-7039-7239

3rd Daniel Seebacher
*Subsequent GmbH*
Konstanz, Germany
0000-0003-0097-5855

4th Prof. Dr.-Ing. Steffen Müller
*Fachgebiet Kraftfahrzeuge*
*TU Berlin*
Berlin, Germany
0000-0002-7831-7695

*Abstract*— Ensuring safe and realistic interactions between automated driving systems and vulnerable road users (VRUs) in urban environments requires advanced testing methodologies. This paper presents a test environment that combines a Vehicle-in-the-Loop (ViL) test bench with a motion laboratory, demonstrating the feasibility of cyber-physical (CP) testing of vehicle-pedestrian and vehicle-cyclist interactions. Building upon previous work focused on pedestrian localization, we further validate a human pose estimation (HPE) approach through a comparative analysis of real-world (RW) and virtual representations of VRUs. The study examines the perception of full-body motion using a commercial monocular camera-based 3D skeletal detection AI. The virtual scene is generated in Unreal Engine 5, where VRUs are animated in real time and projected onto a screen to stimulate the camera. The proposed stimulation technique ensures the correct perspective, enabling realistic vehicle perception. To assess the accuracy and consistency of HPE across RW and CP domains, we analyze the reliability of detections as well as variations in movement trajectories and joint estimation stability. The validation includes dynamic test scenarios where human avatars, both walking and cycling, are monitored under controlled conditions. Our results show a strong alignment in HPE between RW and CP test conditions for stable motion patterns, while notable inaccuracies persist under dynamic movements and occlusions, particularly for complex cyclist postures. These findings contribute to refining CP testing approaches for evaluating next-generation AI-based vehicle perception and to enhancing interaction models of automated vehicles and VRUs in CP environments.

*Keywords*— *Automated Driving, 3D Human Pose Estimation, Motion Capture, Vulnerable Road Users, Vehicle Perception*

## I. Introduction

The integration of automated driving (AD) technologies into urban environments necessitates robust validation procedures, particularly for interactions with vulnerable road users (VRUs) such as pedestrians and cyclists. Conventional test track experiments and numerical simulations remain essential methodologies for ADAS/AD validation; however, they exhibit notable limitations in replicating the complexity, variability, and spontaneity of real-world (RW) human behavior. Scenarios involving potential risks to human participants – such as close-range vehicle-pedestrian encounters – cannot be ethically or safely reproduced on proving grounds, while simulations often lack physical realism and sensory fidelity.

Cyber-physical (CP) environments offer a promising extension to existing validation strategies by enabling safe, repeatable experiments in controlled settings with "ready-to-drive" vehicles and real human subjects. Unlike test track-based setups, a network of test benches – comprising Vehicle-in-the-Loop (ViL), Pedestrian-in-the-Loop (PiL), and Cyclist-in-the-Loop (CiL) – allows for close observation of multi-phase interactions without endangering human participants [1]. These include (i) the vehicle's initial reaction to a VRU, as investigated in [2], (ii) the VRU's behavioral response to the vehicle's motion or presence, as explored in [3] and [4], and (iii) mutual adaptation phenomena that may emerge through the interaction as highlighted in [5], which utilizes a distributed simulation environment to investigate the vehicle-pedestrian interaction. Investigating such processes is essential for the development of accurate behavioral prediction models and the refinement of interaction strategies in AD systems.

To realize this potential, it is crucial to demonstrate that realistic human motion can be captured, interpreted, and recreated in test bench environments in a manner that ensures system perception and behavior remain consistent with RW conditions. In prior work, we already introduced the connected test benches setup and focused on feeding pedestrian motion capture data into the virtual test field (VTF) to examine the impact on vehicle's perception, particularly with respect to skeleton detection, position estimation and latency [1].

In this paper, we further investigate the experimental setup of connected test benches toward establishing the viability of CP testing environments for future VRU protection studies and lay the foundation for more complex investigations into mutual human-vehicle interaction dynamics in safety-critical contexts

## II. Method

To assess the transferability of VRU perception performance between RW and CP environments, a comparative experiment strategy was developed. The objective is to investigate whether the detection behavior of the vehicle under test (VUT) with respect to pedestrians and cyclists is consistent across a real and a virtual test field in combination with ViL and PiL/CiL. This includes analyzing whether similar recognition patterns and misdetections occur in both domains.

The VUT, a real automated shuttle bus, uses a demonstrator "Track and Follow" function to follow a VRU. This results in a closed-loop system where the VUT continuously adapts its motion based on the VRU's trajectory. This setup enables the investigation of the interactions, specifically the vehicle's initial reaction to a VRU, under consistent perceptual conditions.

The VRU's motion in CP environments is controlled via a PiL/CiL test bench, where a human avatar is operated directly using keyboard inputs. Although motion capture of a human test subject is supported by the PiL/CiL setup and could be used instead, the test focuses primarily on vehicle perception, making the simpler input method sufficient. The setup is shown in Fig. 1. Captured data are streamed into the VTF to ensure behaviorally equivalent conditions. By applying identical scenario sequences and aligning viewpoints consistently, this setup enables a systematic comparison of human pose estimation (HPE) outputs between RW and CP domains.

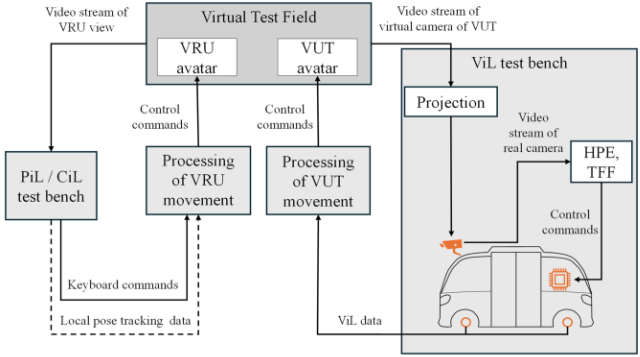

Fig. 1. Experimental CP setup of connected test benches

### A. Test Cases

Three movement patterns of the VRU are used to generate representative interaction constellations: Straight-line movement, lane change maneuver and circular movement. For analysis, these scenarios are decomposed into three perspective categories, which reflect the relative angle and movement direction between the VUT and VRU during the interaction and form the basis for comparing detection consistency: Straight (*S*), Diagonal (*D*) and Crosswise (*C*).

In the RW trials, VUT and VRU move on a predefined track marked by cones to ensure consistent positioning and heading angles across repetitions. The same geometrical layout and VRU behavior are replicated in the CP environment. In both domains, the automated bus operates using the same functional control architecture including real-time, camera-based perception and pose estimation as detailed in Section II.D (Track and Follow Function).

The evaluation compares the distances between the VRU and the VUT based on HPE output. In the RW setup, these are referenced against ground truth data generated by fusing LiDAR and camera-based detections. In CP tests, ground truth positions and orientations of the VRU are directly extracted from the VTF.

By applying the same sequence of scenarios and categorizing perspectives consistently, this strategy enables a structured comparison of HPE under equivalent dynamic conditions. The corresponding test cases are summarized in TABLE I.

TABLE I. TEST CASES

| Test Case No. | VRU | Domain | Perspective |
|---|---|---|---|
| 1-3 | Pedestrian | RW | *S* |
| | | | *D* |
| | | | *C* |
| 4-6 | Pedestrian | CP | *S* |
| | | | *D* |
| | | | *C* |
| 7-9 | Cyclist | RW | *S* |
| | | | *D* |
| | | | *C* |
| 10-12 | Cyclist | CP | *S* |
| | | | *D* |
| | | | *C* |

### B. Definition of Evaluation Metrics

To systematically evaluate the performance of the HPE system under both RW and CP conditions, we define three key metrics: Detection Reliability, Relative Distance Stability, and Joint Stability.

Detection Reliability assesses the consistency of successful person detection throughout a scenario. Two failure modes are considered: *No Detects*, defined as frames in which the HPE system fails to detect a VRU despite their clear visibility in the scene, and *False Detects*, referring to frames where the system erroneously identifies additional persons, although only one VRU is present. These metrics reveal fundamental limitations in the HPE's ability to differentiate VRUs from the scene.

Relative Distance Stability quantifies the smoothness and temporal consistency of the VRU's estimated position relative to the VUT. Specifically, the trajectory of the VRU's hip joint (ID 0) is analyzed within the VUT's coordinate frame to capture relative motion. To isolate local inconsistencies, a smoothed trajectory is generated using a moving mean filter followed by cubic spline smoothing. The local variability is then computed as the absolute deviation between the raw and smoothed position signals over time. High deviations indicate erratic joint estimates and reduce the relative distance stability.

Joint Stability evaluates the consistency of the internal human pose structure. It is assessed by analyzing the temporal standard deviation of each joint's distance relative to the hip anchor joint (ID 0). This local reference accounts for natural movement while minimizing global translation effects. Similar smoothing and variability computations as in relative distance stability are applied to identify instabilities in individual joints. High Joint Stability indicates minimal unintended fluctuations in 3D position estimation of joints.

### C. Human Pose Estimation

For our analysis we utilized a commercial 3D HPE developed by Subsequent GmbH. General functionalities as well es performance metrics during gait analysis in comparison to marker-based motion capture systems are described in [6]. The HPE derives 24 anatomical keypoints of persons from video recordings to model skeletal structures and estimate joint kinematics (see Fig. 2).

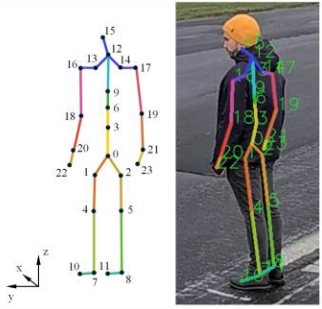

Fig. 2. 3D skeletal structure of HPE oriented in x-direction with 24 keypoints

The process begins with a deep learning-based bounding box detector that identifies individuals in each frame and generates initial bounding boxes. These are used to extract and resize person-specific crops for standardized input. A second-stage convolutional neural network then predicts 24 3D keypoints per person relative to the camera by generating a volumetric heatmap, encoding the probability of a keypoint's 3D location. These 3D heatmaps provide a dense spatial representation, improving depth estimation and aiding joint disambiguation, especially under occlusion or overlap.

The predicted keypoints are assembled into a skeletal graph that models the kinematic relationships between joints, preserving anatomical coherence. To refine the initial estimates, the system applies anatomical priorities that encode biomechanical constraints and typical human movement patterns. These priors guide the pose toward physically plausible configurations, reducing errors from ambiguous or partially occluded input.

The output is a frame-wise reconstruction of 3D skeletal motion, producing spatiotemporal data based on a model trained with approximately one million annotated images containing 24 keypoints. Its statistical modeling and the diversity of the training contribute to stable joint detections across varying conditions and perspective changes.

*D. Track and Follow Function*

As a demonstrator, we examine a simple "Track and Follow" function (TFF) to evaluate object detection and tracking performance. Although the function itself is not the focus of the validation, it enables comparative analysis of 3D HPE consistency, tracking behavior, and closed-loop control performance in both RW and CP environments.

The function follows a modular sense-plan-act control loop and relies exclusively on input from the 3D HPE system. In the sense module, the monocular camera provides image streams processed by the HPE system to extract joint positions of all detected persons. In the plan module, the system continuously selects the closest person within the vehicle's forward field of view (FoV) as the tracking target and detects a start gesture when an arm is raised above head level; the same gesture is used to trigger a stop command. Once activated, a PID controller computes longitudinal velocity commands to maintain a constant distance, while lateral control calculates steering commands to minimize the offset between the VRU and the camera centerline. The act module transmits these controller commands to the actuators via the vehicle's CAN bus, safeguarded by a drive-by-wire interface with conditional override logic for a system operator.

## III. Experimental Setup

The RW experiments were conducted using a Navya autonomous shuttle bus equipped with a monocular front-facing First Sensor DC3K-1 camera and a Velodyne VLP-16 LiDAR for environment perception. The shuttle was augmented with a dedicated measurement PC running a ROS2-based data acquisition framework, which synchronized and logged inputs from vehicle-internal CAN-bus, camera, LiDAR, the 3D HPE system, and TFF. This setup enabled real-time processing of sensor data, control commands, and pose estimations in outdoor conditions. The test track is located on an airstrip featuring slight inclination and surface irregularities, introducing moderate RW disturbances.

The experimental CP setup (see Fig. 1**Fehler! Verweisquelle konnte nicht gefunden werden.**) builds upon our prior work in [1], extending it through the integration of the TFF and an enhanced VTF. All components are connected via a synchronized, cross-domain ROS2 network that ensures time-aligned data exchange and logging across subsystems. The network captures multimodal inputs and outputs, including CAN-bus data, camera images from the VUT, 3D joint positions from the HPE system, TFF control signals, and ground-truth position and orientation data from the VTF.

The CP setup enables interactive experiments where virtual and physical components influence each other, while ensuring full safety for human participants.

*A. Virtual Test Field*

The VTF is implemented using Unreal Engine version 5.4, enabling real-time rendering of all simulation views from synchronized simulation states. These views are available on the respective test benches to maintain temporal coherence across the system.

The software implementation is structured as a multiplayer application, wherein each test bench connects as a client and, based on its type, assumes a predefined functional role within the scenario. The virtual environment is a digital reconstruction of the RW test track, developed as a "level" within Unreal Engine. This setup ensures consistency between RW and CP test domains.

Visual models from Epic Games MetaHumans [7] are employed as VRU representations to mitigate rendering issues observed in earlier work [1], particularly those related to reflective metallic surfaces on standard Unreal Engine human avatars.

*B. Vehicle-in-the-Loop Test Bench*

The real automated vehicle on the ViL test bench responds to virtual scenarios by steering, accelerating, and braking against physically coupled actuators that simulate realistic force and torque feedback. Electrical wheel motors apply resistance torques to emulate longitudinal driving dynamics, while a linear steering actuator (Steering Force Emulator) generates counterforces at the steering rack, enabling realistic yaw behavior under test bench conditions [8].

A real-time driving dynamics simulation processes the vehicle's control signals – such as steering angle and applied torques – to continuously compute the dynamic state of the vehicle (e.g., velocity, wheel speeds, and orientation). This state information is used to update the position and orientation of a vehicle avatar within the shared VTF, ensuring consistent behavior between the physical and virtual domains. Further technical details and evaluations of the ViL test bench can be found in our previous work [8, 9].

*C. Camera Stimulation*

A key component of the ViL setup is a projection-based camera stimulation technique, ensuring consistent visual input for the vehicle's perception system in both RW and CP domains. As before, we use VUT's camera (1280 x 960 px, 20 fps, 7.6° pitch) and feed the images to an instance of the 3D HPE.

The projection system must be calibrated such that the virtual scene captured by the virtual camera aligns precisely with the image plane of the physical camera. This requires that the horizontal FoV of the virtual camera corresponds to the width of the projected image. Furthermore, the resolution of the virtual camera must match the native resolution of the projector to ensure the correct aspect ratio.

Fig. 3 illustrates the setup along with the respective image planes of physical and virtual camera. Due to the geometric constraints of the shuttle bus and the steep projection angle required for exterior mounting, the projector is installed inside the vehicle. It allows a sufficiently large image to be displayed on the windshield. While no significant optical distortion observed, minor deviations due to the windshield glass cannot be entirely excluded.

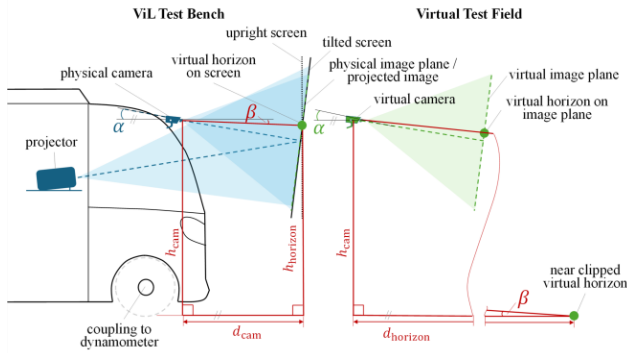

Fig. 3. Camera Stimulation Setup

As the projector and camera are not positioned at the same height (z-axis) and distance (x-axis), these geometric deviations must be compensated both physically (through positioning and orientation of the projector) and digitally (via keystone correction). The goal is to ensure centered and orthogonal image alignment respectively to the physical environment.

The horizontal field of view $FoV_h$ of the virtual camera can be derived from width $W$ of the projected image and distance $d_{cam}$ between the camera and the projection surface by Eq. (1). The vertical FoV results from the aspect ratio of the virtual camera.

$$FoV_h = 2 \arctan\left(\frac{W}{2\,d_{cam}}\right) \quad (1)$$

Due to the height difference between projector and camera, the vertical alignment must also be adapted. A clipped horizon is rendered at a finite distance $d_{horizon}$ from the virtual camera. Due to this finite distance, the horizon does not lie at the vertical center of the image. Moreover, it is further shifted depending on the camera pitch α. The physical height $h_{horizon}$ on screen can be calculated using the intercept theorem, as shown in Eq. (2), and adjusted by modifying the projector's pitch angle. Tilting the projector may require subsequent keystone correction.

$$h_{horizon} = \frac{h_{cam}(d_{horizon} - d_{cam})}{d_{horizon}} \quad (2)$$

To further compensate for the camera pitch, the projection surface can be physically tilted so that the camera is oriented perpendicularly to the screen. If the projection angle is nearly orthogonal and the camera pitch is minimal, the distortion of an image projected to an upright screen can be considered negligible, assuming the small-angle approximation holds and the deviations remain within the limits of the system calibration.

To verify the accuracy of the camera stimulation, a qualitative comparison between the RW and CP perspectives captured both by the real camera were conducted. Two rectified images were superimposed in Fig. 4, with added semi-transparent appearance for better optical comparability.

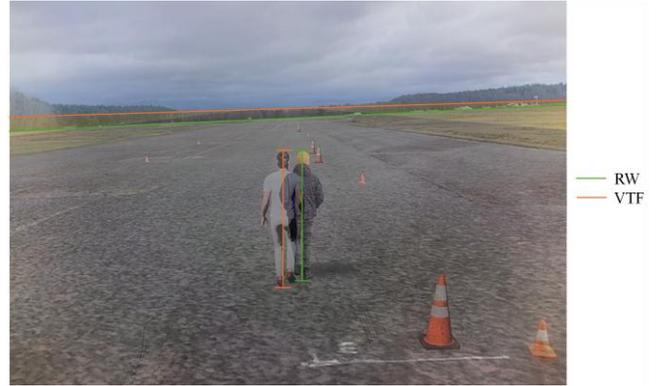

Fig. 4. Perspective comparison of RW and VTF at 10 m relative distance of pedestrian during test cases 1 and 4

A walking pedestrian served as a reference; the frame selected for comparison corresponds to the moment when the pedestrian is located 10 m in front of the camera in both domains. In the CP scene, this position was verified using ground truth data from VTF; in the RW scene, it was determined through sensor fusion of camera and LiDAR data. The pedestrian appears 1.2% taller in the CP image, which is acceptable given the calibration accuracy limits. While the heads are aligned at identical height and both horizons appear equally tilted around the vehicle's x-axis due to the cameras roll angle, the horizon line in the RW image appears lower – likely caused by terrain elevation and surrounding vegetation. As the visual horizon's actual depth cannot be reliably estimated, the road surface was used as a more consistent reference for image alignment.

## IV. EXPERIMENTAL RESULTS

The following results are structured along the evaluation metrics defined in Section II.B, addressing detection reliability, relative distance stability, and joint stability across RW and CP domains.

### A. Detection Reliability

Across all evaluated cases, no *No Detects* were observed in either the CP or RW environment, suggesting a high baseline reliability of the HPE system under the tested conditions.

However, *False Detects* occurred in the CP environment in two cases, specifically for Perspective *D*. In these instances, the virtual cone object near the VRU was incorrectly detected as an additional person. The distinction between the real and virtual cone is illustrated in Fig. 4, which highlights a significant difference in height. This effect was not observed in the corresponding RW recordings, indicating that specific characteristics of the virtual environment can introduce detection artifacts that do not occur in the real world.

### B. Relative Distance Stability from Straight Perspective

As highlighted in our previous work [1], the stability of 3D position estimation via HPE decreases with increasing relative distance between VUT and VRU, with fluctuations of more than 1 m were observed at around 16 m. In the present study, we reassess this effect using an updated, more realistic VRU avatar model in the CP environment and compare the results to measurements collected on the RW test track.

Fig. 5 (top row) shows the relative longitudinal distances between VUT and VRU – pedestrian and cyclist respectively – for both CP (left) and RW (right) environments, covering a range from 5 m to 24 m. The corresponding local variability is plotted in the lower row.

The analysis confirms a consistent trend: higher relative distances are associated with increased variability in position estimation. At estimated 20 m, local deviations reach up to 0.16 m in CP and 0.16 m in RW for pedestrians, while cyclist detections show 0.23 m variability in CP and significantly higher 0.43 m in RW. This indicates that while the improved avatar enables comparable stability for pedestrian detection across domains, cyclist detection remains more sensitive to RW influences such as motion blur or occlusion.

### C. Joint Stability from Various Perspectives

To assess the stability of HPE across different environments, we analyze the distances of 23 detected joints relative to the hip anchor point (ID 0). The motion trajectories of these joints vary depending on human movement, with distal joints such as hands and feet exhibiting greater variability, while central joints like the torso remain comparatively stable.

For each joint, we calculate the standard deviation *SD* of its relative distance to the anchor point over time for pedestrian (1-6) and cyclist (7-12) test cases, where both RW and CP test cases are compared for each perspective. Subsequently, the relative error *RE* between RW and CP is determined as:

$$RE = \left|\frac{SD_{RW} - SD_{CP}}{SD_{CP}}\right| \qquad (3)$$

This comparison provides insights into how well joint stability observed in RW is replicated in the virtual environment, particularly under dynamic motion conditions. By focusing on local joint variability relative to a fixed reference, we account for both movement amplitude and potential distortions in pose estimation across domains.

The results for the hand (ID 22, 23), foot (ID 7, 8), and shoulder (ID 16, 17) joints are summarized in TABLE II (Pedestrian) and TABLE III (Cyclist), showing *SD* and *RE* values.

TABLE II. JOINT DETECTION STABILITY OF PEDESTRIAN

| Joint ID | Perspective *S* | | | Perspective *D* | | | Perspective *C* | | |
|---|---|---|---|---|---|---|---|---|---|
| | RW | CP | *RE* | RW | CP | *RE* | RW | CP | *RE* |
| | *SD* in cm | | in % | *SD* in cm | | in % | *SD* in cm | | in % |
| 22 | 1.55 | 2.38 | 53.4 | 1.32 | 2.15 | 63.1 | 2.05 | 2.87 | 39.8 |
| 23 | 1.56 | 2.44 | 56.5 | 1.29 | 1.37 | 6.5 | 1.33 | 2.26 | 70.1 |
| 7 | 2.54 | 1.95 | 23.4 | 2.15 | 1.61 | 25.1 | 2.28 | 1.73 | 24.3 |
| 8 | 2.32 | 2.12 | 8.6 | 2.87 | 1.9 | 33.9 | 3.01 | 2.15 | 28.5 |
| 16 | 0.44 | 0.89 | 103.5 | 0.49 | 0.38 | 21.8 | 0.38 | 0.55 | 44.1 |
| 17 | 0.36 | 0.67 | 86.6 | 0.58 | 0.39 | 33 | 0.39 | 0.63 | 60.5 |

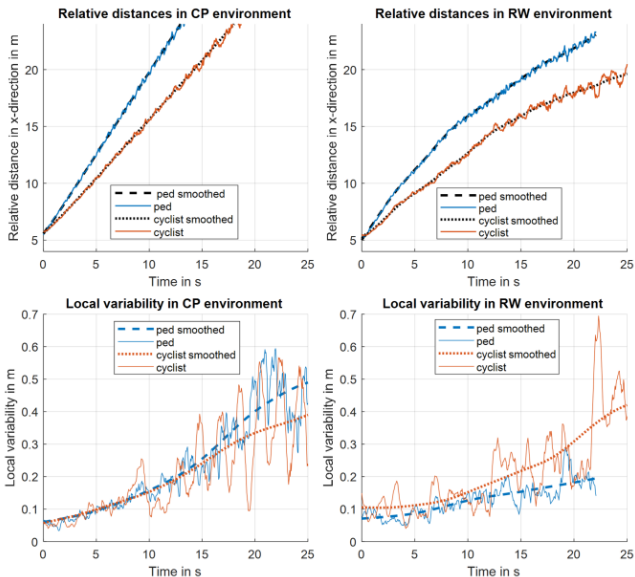

Fig. 5. Stability analysis of HPE in CP and RW environment

TABLE III. JOINT DETECTION STABILITY OF CYCLIST

| Joint ID | Perspective *S* | | | Perspective *D* | | | Perspective *C* | | |
|---|---|---|---|---|---|---|---|---|---|
| | RW | CP | *RE* | RW | CP | *RE* | RW | CP | *RE* |
| | *SD* in cm | | in % | *SD* in cm | | in % | *SD* in cm | | in % |
| 22 | 1.75 | 0.65 | 62.9 | 2.2 | 1.29 | 41.5 | 1.21 | 2.10 | 73.2 |
| 23 | 1.63 | 0.76 | 53.0 | 2.64 | 1.16 | 56.1 | 1.3 | 1.43 | 10.4 |

| Joint ID | Perspective S | | | Perspective D | | | Perspective C | | |
|---|---|---|---|---|---|---|---|---|---|
| | RW SD in cm | CP SD in cm | RE in % | RW SD in cm | CP SD in cm | RE in % | RW SD in cm | CP SD in cm | RE in % |
| 7 | 3.79 | 3.3 | 13.0 | 3.82 | 3.26 | 14.6 | 4.84 | 2.96 | 38.9 |
| 8 | 3.76 | 3.93 | 4.3 | 3.54 | 2.16 | 38.9 | 3.95 | 1.81 | 54.2 |
| 16 | 0.55 | 0.35 | 36.7 | 0.85 | 0.39 | 53.9 | 0.38 | 0.47 | 23.6 |
| 17 | 0.46 | 0.32 | 30.7 | 0.94 | 0.33 | 64.9 | 1.02 | 0.38 | 62.6 |

For the pedestrian cases, high *RE* values in Perspective *S*, particularly in the shoulder and arm joints, can be attributed to a sudden rotational shift of the upper body detected by the HPE in the CP domain. Although the VRU isn't turning, the HPE erroneously assumes a lateral twist near the end of the trial. In Perspective *D*, joint detection is stable and comparable between CP and RW, resulting in low *RE* values. In Perspective *C*, the RW data reveal that the pedestrian bent their left arm during the trial to check a wristwatch for speed monitoring. This gesture led to a distinctly lower *SD* for the left arm compared to the right, which is also reflected in the corresponding *RE* values when compared to the CP environment, where no such arm movement occurred.

The RW cyclist tests show pronounced steering corrections and unstable riding behavior due to low speed of ca. 6 km/h, affecting the stability of the hand and shoulder joints. These dynamic movements have not yet been replicated in the CP environment but are planned for future updates. In Perspectives *D* and *C*, occlusion effects are evident: In side views, the HPE struggles with detecting the occluded right shoulder and foot, leading to increased instability. Interestingly, higher variability is observed in RW compared to CP, suggesting that RW occlusions pose greater challenges to consistent joint detection.

## V. Summary and Outlook

In this study, we evaluated the detection stability and reliability HPE across RW and CP environments. The analysis focused on three main aspects: Detection reliability, stability over varying distances, and joint stability based on relative distances to a hip anchor point.

Regarding detection reliability, no missing detections were observed in either domain. However, false detections occurred in the CP environment. With respect to detection stability, the results confirmed that relative distance significantly influences the local variability of HPE, consistently across both domains. Pedestrian detection demonstrated comparable stability between CP and RW, whereas cyclist detection was more sensitive to RW-specific conditions such as motion blur and occlusions. The evaluation of joint stability showed that pedaling movements and steering adjustments to maintain balance led to higher variability on the test track, particularly in side views. Although the CP environment successfully replicated RW variability for stable motion patterns, it exhibited limitations in representing dynamic behaviors and occlusion effects.

Based on the identified limitations, the following improvements are recommended for future investigations:

- The CiL should enable the integration of dynamic VRU behaviors at low speeds, such as steering corrections and unstable riding patterns, to more accurately replicate RW variability.
- The pedestrian avatar's upper body should be refined to prevent unnatural rotational detections.
- Non-human objects in the CP scene should be optimized to reduce false detections. This could be achieved by visually distinguishing static objects from VRUs through adjusted textures, shapes or the use of dynamic semantic tagging.

ACKNOWLEDGMENT

This work was partly funded by the German Federal Ministry for Economic Affairs and Climate Action (BMWK) and partly financed by the European Union in the frame of NextGenerationEU within the project "Solutions and Technologies for Automated Driving in Town" [10] (grant FKZ 19A22006P).

Parts of the text were edited with the assistance of ChatGPT (OpenAI) for language and phrasing. All scientific content was created and verified by the authors.